\documentclass[reprint, amsmath, amssymb, aps, pre, floatfix]{revtex4-2}
\usepackage{float}
\makeatletter
\let\newfloat\newfloat@ltx
\makeatother
\usepackage{graphicx}
\usepackage[utf8]{inputenc}
\usepackage{xcolor}
\usepackage{subfigure}
\usepackage{romannum}
\usepackage{url}
\usepackage{soul}

\makeatletter
\DeclareFontFamily{OMX}{MnSymbolE}{}
\DeclareSymbolFont{MnLargeSymbols}{OMX}{MnSymbolE}{m}{n}
\SetSymbolFont{MnLargeSymbols}{bold}{OMX}{MnSymbolE}{b}{n}
\DeclareFontShape{OMX}{MnSymbolE}{m}{n}{
    <-6>  MnSymbolE5
   <6-7>  MnSymbolE6
   <7-8>  MnSymbolE7
   <8-9>  MnSymbolE8
   <9-10> MnSymbolE9
  <10-12> MnSymbolE10
  <12->   MnSymbolE12
}{}
\DeclareFontShape{OMX}{MnSymbolE}{b}{n}{
    <-6>  MnSymbolE-Bold5
   <6-7>  MnSymbolE-Bold6
   <7-8>  MnSymbolE-Bold7
   <8-9>  MnSymbolE-Bold8
   <9-10> MnSymbolE-Bold9
  <10-12> MnSymbolE-Bold10
  <12->   MnSymbolE-Bold12
}{}

\let\llangle\@undefined
\let\rrangle\@undefined
\DeclareMathDelimiter{\llangle}{\mathopen}%
                     {MnLargeSymbols}{'164}{MnLargeSymbols}{'164}
\DeclareMathDelimiter{\rrangle}{\mathclose}%
                     {MnLargeSymbols}{'171}{MnLargeSymbols}{'171}
\makeatother

\usepackage{float}
\makeatletter
\let\newfloat\newfloat@ltx
\makeatother
\usepackage{algorithm} 
\usepackage{algpseudocode}

\definecolor{darkblue}{rgb}{0,0,0.6}
\definecolor{darkred}{rgb}{0.6,0,0}
\definecolor{darkgreen}{rgb}{0,0.6,0}

\usepackage[colorlinks=true,urlcolor=darkblue,citecolor=darkblue,linkcolor=darkred,hyperfootnotes=false]{hyperref}

\AtBeginDocument{\pagenumbering{arabic}}
\begin{document}

\title{Exploring how deep learning decodes anomalous diffusion via Grad-CAM}

\author{Jaeyong Bae}
\affiliation{Department of Physics, Korea Advanced Institute of Science and Technology, Daejeon 34141, Korea}

\author{Yongjoo Baek}
\email{y.baek@snu.ac.kr}
\affiliation{Department of Physics and Astronomy \& Center for Theoretical Physics, Seoul National University, Seoul 08826, Korea}

\author{Hawoong Jeong}
\email{hjeong@kaist.edu}
\affiliation{Department of Physics, Korea Advanced Institute of Science and Technology, Daejeon 34141, Korea}
\affiliation{Center of Complex Systems, Korea Advanced Institute of Science and Technology, Daejeon 34141, Korea}

\begin{abstract}
While deep learning has been successfully applied to the data-driven classification of anomalous diffusion mechanisms, how the algorithm achieves the feat still remains a mystery. In this study, we use a well-known technique aimed at achieving {\em explainable AI}, namely the Gradient-weighted Class Activation Map (Grad-CAM), to investigate how deep learning (implemented by ResNets) recognizes the distinctive features of a particular anomalous diffusion model from the raw trajectory data. Our results show that Grad-CAM reveals the portions of the trajectory that hold crucial information about the underlying mechanism of anomalous diffusion, which can be utilized to enhance the robustness of the trained classifier against the measurement noise. Moreover, we observe that deep learning distills unique statistical characteristics of different diffusion mechanisms at various spatiotemporal scales, with larger-scale (smaller-scale) features identified at higher (lower) layers.

\end{abstract}

\maketitle

\section{Introduction}\label{sec:Intro}


As our ability to generate, store, and analyze large datasets has dramatically increased, data analysis has become an essential component of modern science, including physics~\cite{Mehta2019}. Recent developments of deep learning have revolutionized the way we extract information from the data, allowing us to recognize meaningful patterns from complex, unprocessed empirical data~\cite{Carleo2019}. Besides applications to speech recognition~\cite{speechrev}, computer vision~\cite{cvrev}, self-driving cars~\cite{selfdriv}, and natural language processing~\cite{gpt}, deep learning has demonstrated remarkable efficacy for analyzing various physical data, such as tracking specific events in complex environments~\cite{atlas}, extracting signals from astronomical data~\cite{gravwave}, and predicting valid protein configurations~\cite{alphafold}.

More recently, deep learning was also applied to analyze anomalous diffusion. The phenomenon, characterized by nonlinear growth in time of the mean square displacement, is observed in diverse disciplines encompassing biology~\cite{biology1, biology2, biology3}, social science~\cite{social1}, and finance~\cite{stock1}. Various mathematical models have been proposed to describe the mechanism of anomalous diffusion, including continuous-time random walk~\cite{CTRW1}, fractional Brownian motion~\cite{FBM1}, L\'{e}vy walk~\cite{LW1}, annealed transient time motion~\cite{ATTM1}, and scaled Brownian motion~\cite{SBM1}. However, identifying the correct mechanism underlying a given trajectory is a challenging task, as long-range temporal correlations often make it difficult to identify useful statistical features~\cite{anomalousdiffusion_property}. Despite various statistical methods proposed thus far~\cite{stats1, stats2, stats3, stats4, stats5}, a consensus on this issue is yet to be reached. To address this long-standing problem, the recent anomalous diffusion (AnDi) challenge applied a variety of machine learning techniques to the task~\cite{andi}, confirming the enhanced accuracy and efficiency of deep-learning approaches~\cite{ml1, ml2, ml4, ml5, ml6, ml7, ml8} compared to the conventional statistical methods.

However, the black-box nature of deep learning poses a significant challenge as to how to interpret its outcomes---it is unclear which features of the data are used by the algorithm to perform the given task. Notably, a recent study~\cite{ml4} employed Bayesian learning techniques to provide error estimates of the outcomes, whose behaviors as the training data are varied indicate how properties of the underlying diffusion mechanism affect the learning performance. However, this approach does not explicitly reveal which features of the data lead to the observed outcomes.

In this study, we propose a method to highlight regions within the input trajectory that are key to how deep learning classifies the diffusion mechanisms. Similar problems have been addressed in different contexts, especially computer vision, using the techniques aiming to achieve \textit{explainable AI}~\cite{xai1, xai2, xai3, xai4, xai5, xai6, xai7, cam, gradcam, gradcam++, gradcam1, gradcam2, gradcam3}. Among these, noting that many of the deep learning methods in the AnDi challenge were based on convolutional neural networks (CNNs), we employ the \textit{gradient-weighted class activation mapping} (Grad-CAM) developed for the architecture. The versatility of Grad-CAM has been demonstrated in various problems, which is now widely accepted as a standard technique of explainable AI~\cite{cam, gradcam, gradcam++, gradcam1, gradcam2, gradcam3}. By integrating Grad-CAM with the diffusion model classifier based on deep learning, we aim to identify substructures within trajectories that contain key information about the underlying mechanism, which can be further applied to enhancing the robustness of the learning performance.


The rest of this paper is organized as follows. In Sec.~\ref{sec:methods}, we define the task of classifying particle trajectories according to the underlying anomalous diffusion mechanism and describe the deep learning method that performs the task. In Sec.~\ref{sec:relevance}, we introduce Grad-CAM and demonstrate its relevance to meaningful features of the data. In Sec.~\ref{sec:statistics}, we identify statistical quantities that may be useful for the classification task and discuss how they are correlated with the Grad-CAM outcomes. Finally, we summarize and conclude in Sec.~\ref{sec:summary}.

\section{Classification task} \label{sec:methods}
\begin{figure*}
\centerline{\includegraphics[width=\textwidth]{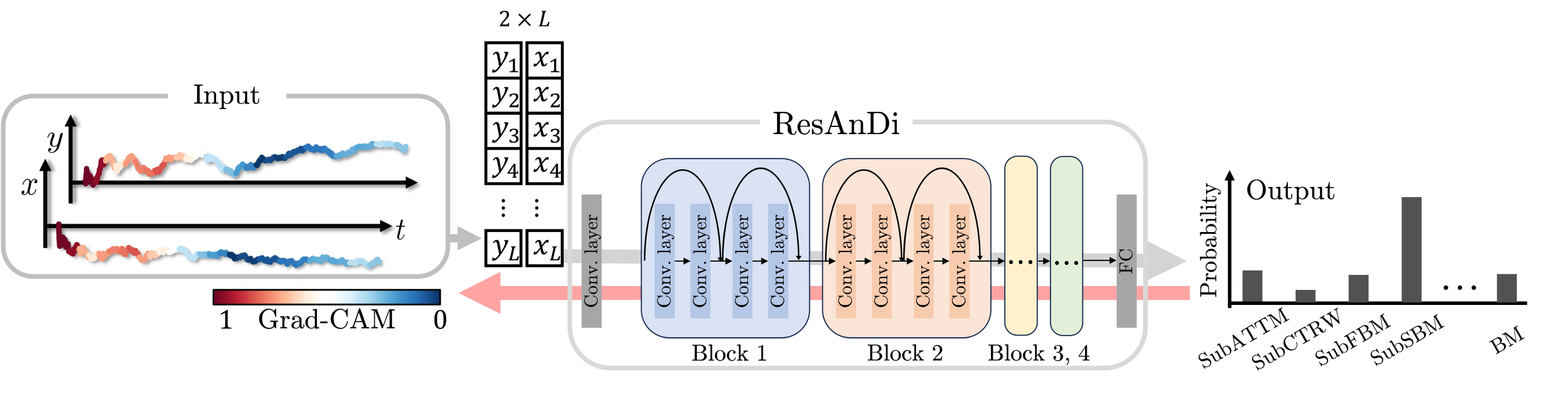}}
\caption{Schematic illustration of model classification via ResAnDi and evaluation of the Grad-CAM score. Processing a time series representing a two-dimensional particle trajectory via $18$ layers, ResAnDi yields a vector whose components indicate the probabilities that the trajectory belongs to each of the eight classes of diffusion mechanisms described in the main text. Moreover, by calculating which nodes of the last convolutional layer contribute more to correct classification, the Grad-CAM score is assigned to each subinterval of the trajectory.
}
\label{figure1}
\end{figure*}

Before proceeding, let us clarify what we aim to accomplish with deep learning. The task is to identify the anomalous diffusion mechanism underlying a two-dimensional particle trajectory. The following describes how we generate the particle trajectories, which deep learning algorithm we employ, and how effectively the algorithm performs the task.

\subsection{Trajectory generation} \label{ssec:traj_gen}

Trajectories exhibiting anomalous diffusion are generated using the Python package provided by the AnDi challenge~\cite{andi,andiDataset}. This package encompasses five standard models of anomalous diffusion: annealed transient time motion (ATTM), continuous-time random walk (CTRW), fractional Brownian motion (FBM), L\'{e}vy walk (LW), and scaled Brownian motion (SBM). While the original AnDi challenge focused on classifying trajectories into these five categories, we also require the neural network to distinguish between subdiffusive and superdiffusive trajectories. By adding the prefixes ``Sub-" and ``Sup-" to indicate subdiffusion and superdiffusion, respectively, our classification task involves seven distinct mechanisms of anomalous diffusion: SubATTM, SubCTRW, SubFBM, SubSBM, SupFBM, SupLW, and SupSBM. Additionally, we include ordinary Brownian motion (BM) as the eighth mechanism.

A dataset, be it for training, validation, or testing, comprises an equal number of trajectories for each of the eight mechanisms. For models exhibiting subdiffusion, the diffusion exponents of the corresponding trajectories are uniformly distributed in the interval $[0.1, 0.9]$. Similarly, for models exhibiting superdiffusion, the diffusion exponents are uniformly distributed in the interval $[1.1, 1.9]$. Each trajectory is rescaled so that the displacement per unit time ($\Delta t = 1$) has unit variance. Unless specified otherwise, the temporal duration of each trajectory is uniformly distributed between $10$ and $1000$.

See the Supplementary Information~\cite{supple} for more details regarding the definition of each anomalous diffusion model and the preparation of datasets.

\subsection{Deep learning algorithm} \label{ssec:deep_learning}

We introduce ResAnDi, a deep learning algorithm designed to identify the anomalous diffusion mechanism underlying an empirical trajectory. This algorithm utilizes a neural network architecture based on the residual neural network (ResNet)~\cite{resnet}. ResNet enhances the standard CNN by incorporating skip connections between layers, allowing the use of deeper networks while mitigating the vanishing gradient problem. Specifically, our model is based on ResNet18, which consists of one convolutional layer, four convolutional blocks ($4$ convolutional layers with skip connections in each block), and one fully-connected (FC) layer. While ResNet was originally developed for processing RGB images through three channels of two-dimensional arrays, we have modified the architecture to handle time series data of two-dimensional particle trajectories using two channels of one-dimensional arrays. In the end, the network produces a vector with components representing the probabilities that the trajectory belongs to each of the eight mechanisms previously mentioned. See Fig.~\ref{figure1} for a schematic illustration.

To train ResAnDi, we use the PyTorch package~\cite{pytorch} with the categorical cross-entropy loss function and the Adam optimizer~\cite{adam}, employing an early-stopping method to avoid overfitting.
We tested the performance of ResAnDi using a dataset composed of $10^4$ trajectories belonging to each of the eight classes. ResAnDi achieves an overall classification accuracy of $90.36 \%$, which is comparable to the best algorithm~\cite{ml5} submitted to the AnDi challenge~\cite{andi} (which achieved $89.16~\%$), even though our task involves a greater number of classes.

See the Supplementary Information~\cite{supple} for more details regarding the neural network architecture, training procedure, and classification accuracy for each class.

\section{Relevance of Grad-CAM} \label{sec:relevance}

Previous studies have shown that Grad-CAM effectively highlights the most relevant features of an image that contribute to its correct classification~\cite{cam,gradcam}. We show that the same is true for identifying diffusion mechanisms. For this purpose, we employ two different approaches. First, we show that the accuracy of ResAnDi is more adversely affected by targeted erasure of particle trajectories with a higher Grad-CAM score. Second, we show that the accuracy of ResAnDi is more robust against noisy input if the training data are augmented using trajectories with a higher Grad-CAM score.

\subsection{Grad-CAM} \label{ssec:grad_cam}

For completeness, we first give a brief description of Grad-CAM. The method was developed with the architecture of CNNs in mind. It focuses on the last convolutional layer, which is expected to possess the highest-level semantic information while retaining some amount of spatial information~\cite{cnnrepresentation1,cnnrepresentation2,gradcam}. Within this layer, let $A_i^k$ denote the activation of node $i$ of the $k$-th feature map $\mathbf{A}^k$. If the probability of the CNN correctly classifying the input data is $p$, then the influence of the $k$-th feature map on the decision can be quantified as
\begin{align}
a^k \equiv \frac{1}{|\mathbf{A}^k|}\sum_i\frac{\partial p}{\partial A^k_i},
\end{align}
where $|\mathbf{A}^k|$ is the size of the $k$-th feature map.

The Grad-CAM scores $\mathbf{G}$ are obtained by averaging over the activations of all feature maps, where each feature map is weighted by its influence on the correct classification. This is expressed by the formula
\begin{align}
G_i \equiv \sum_k a^k A^k_i.
\end{align}
Given a trained CNN and an input sample, $\mathbf{G}$ can be obtained readily using the standard backpropagation method. In this study, we used the Captum Python package~\cite{captum} to implement the calculation.

It should be noted that the Grad-CAM scores defined above are assigned to the nodes in the last convolutional layer, whose activation pattern is a coarse-grained representation of the original input. To reassign these scores to the nodes of the input data, an interpolation scheme is needed. For efficiency, we partition the input nodes into (approximately) equal-length subintervals, with the number of subintervals matching the number of nodes in the final convolutional layer. This creates a one-to-one correspondence between the nodes in the final convolutional layer and the subintervals in the input layer. Through this correspondence, the Grad-CAM score is transferred from the final convolutional layer to the input trajectory.

\subsection{Targeted erasure based on Grad-CAM} \label{ssec:targeted_erasure}

\begin{figure}
\centerline{\includegraphics[width=0.8\linewidth]{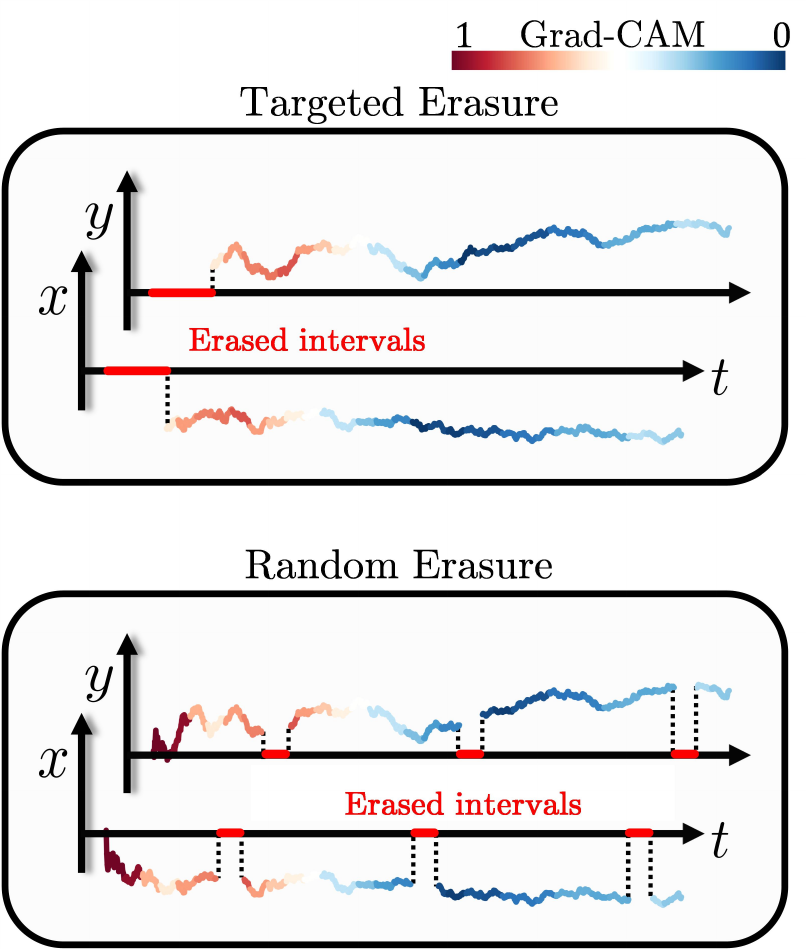}}
\caption{
Examples of particle trajectories in the $xy$-plane whose subintervals are erased (top) by targeting the top $10\%$ of the Grad-CAM score (indicated by the color scale) or (bottom) by random choice.
}
\label{figure2}
\end{figure}

To check whether Grad-CAM captures the relevant features of trajectories, we propose the following test. First, we prepare a new set of trajectories not used in the training, whose temporal duration varies from $10$ to $1000$. Given this test dataset, targeted erasure is implemented as follows. We start with partitioning each trajectory into subintervals so that the Grad-CAM score can be assigned to each of them according to the previously described procedure. By letting ResAnDi classify the trajectory, we obtain the Grad-CAM score of each subinterval. This allows us to modify a trajectory by ``erasing'' all subintervals whose Grad-CAM score falls within a targeted range. More specifically, erasing a subinterval means that the particle is forced to stay at the origin ($x = y = 0$) during the whole subinterval. By comparing the effects of targeted erasure with those of random erasure, where randomly chosen subintervals are erased, we can assess how the Grad-CAM score relates to the presence of useful information about diffusion mechanisms. See Fig.~\ref{figure2} for illustrations of targeted and random erasures.

\begin{figure}
\centerline{\includegraphics[width=\linewidth]{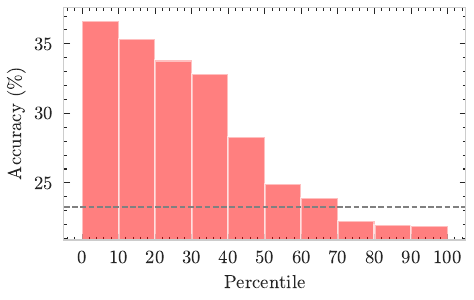}}
\caption{
Classification accuracy of ResAnDi after targeted erasure of subintervals corresponding to each decile of the Grad-CAM score. Removing subintervals with a higher Grad-CAM score results in lower accuracy. For comparison, the effect of random erasure is also shown by a dashed line.
}
\label{figure3}
\end{figure}

In Fig.~\ref{figure3}, we show how the classification accuracy of the ResAnDi is affected by targeted erasure of subintervals whose Grad-CAM scores fall between each consecutive pair of deciles (\textit{i.e.}, every tenth percentile). Clearly, erasing subintervals of the higher Grad-CAM score leads to the lower accuracy of the ResAnDi. Compared to the random erasure of subintervals, erasing subintervals whose Grad-CAM score belongs to the upper $70\%$ has more adverse effects on the classifier. These suggest that certain parts of the particle trajectories contain more information about the underlying diffusion mechanism than others, and that the Grad-CAM score can be used as an indicator of those crucial parts.

\begin{figure}
\centerline{\includegraphics[width=0.8\linewidth]{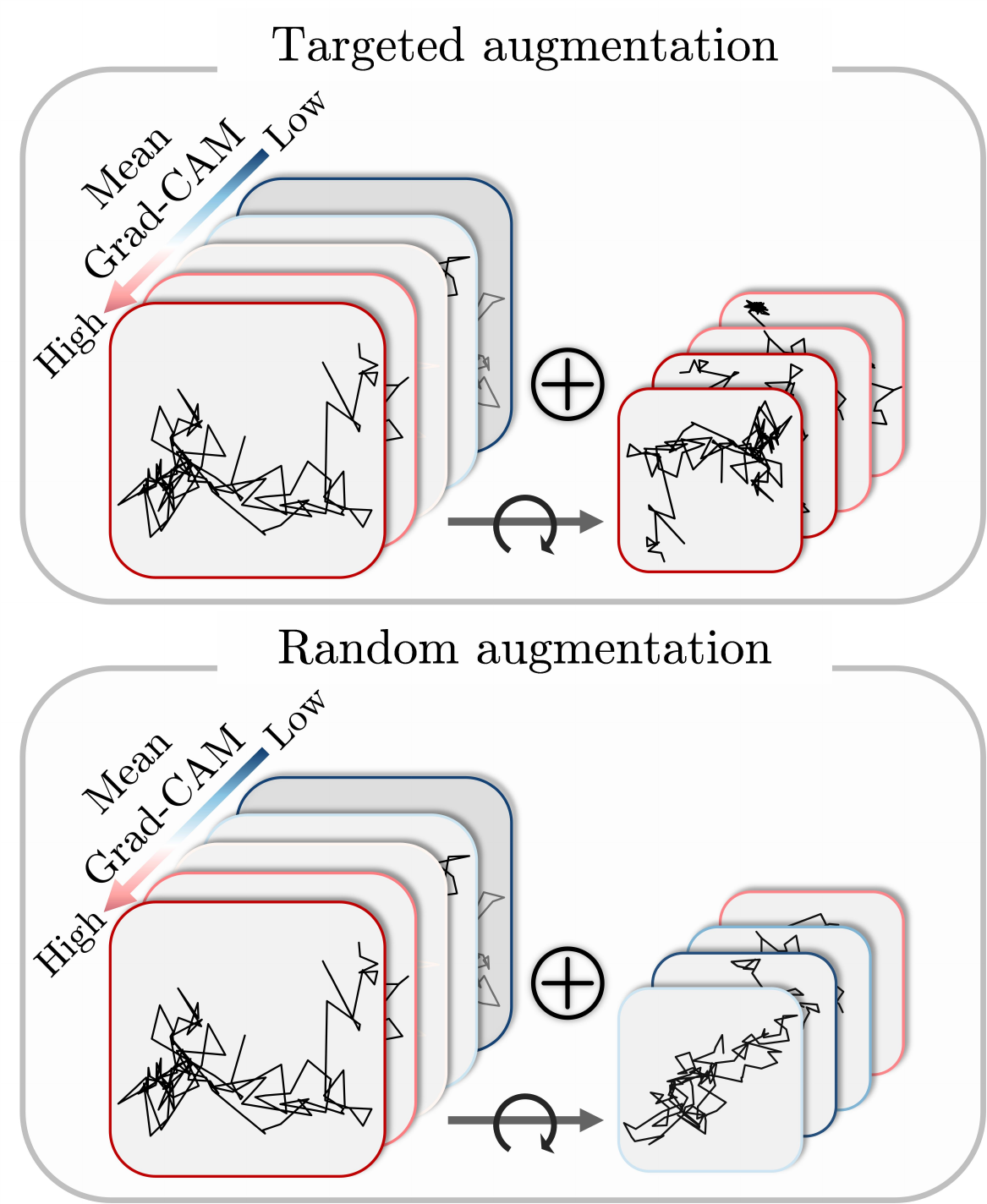}}
\caption{Schematic illustrations of dataset augmentation method. (Top) Using targeted augmentation, trajectories with high a mean Grad-CAM score are rotated by random angles to build the augmented dataset. (Bottom) Using random augmentation, the trajectories to be rotated are chosen at random.
}
\label{figure4}
\end{figure}

\begin{figure}
\centerline{\includegraphics[width=\linewidth]{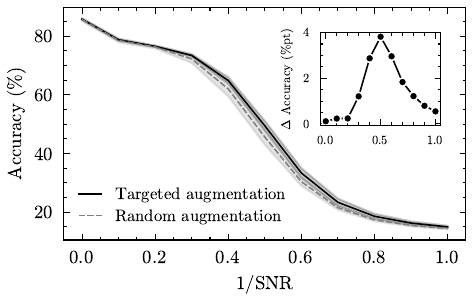}}
\caption{Effects of noise in the unseen test data on the classification accuracy. Targeted augmentation using trajectories belonging to the top $60\%$ of the mean Grad-CAM score exhibits more robust performance against increased noise level. The statistics are obtained from $5$ models trained using each augmentation scheme, with the standard errors indicated by shaded regions. (Inset) Enhanced accuracy due to the use of targeted augmentation.
}
\label{figure5}
\end{figure}

\subsection{Dataset augmentation via Grad-CAM}

Dataset augmentation aims to enhance machine learning performance by expanding the diversity of the training data. This typically involves making variants of certain samples through random cropping, resizing, or rotating, which modifies non-critical aspects of the data while preserving the features crucial for classification tasks~\cite{augsurvey}. However, these conventional approaches do not evaluate the usefulness of individual samples for training, generating an augmented dataset only through random selection. This prompts the question of whether the procedure could be improved by making informed decisions about which samples to augment.

We hypothesize that the Grad-CAM score, which quantifies information on the underlying diffusion mechanism, might also suggest which trajectories are optimal for creating an augmented dataset. To test this, we propose targeted augmentation, which is implemented as follows. First, we select trajectories that ranked in the top $60\%$ based on their mean Grad-CAM scores, calculated by averaging the scores across subintervals of each trajectory. Then, these selected trajectories are rotated by a random angle and added to the original training dataset to form the augmented dataset. See Fig.~\ref{figure4} for a schematic illustration of targeted augmentation as opposed to the conventional random augmentation. Additional details on this procedure can be found in the Supplementary Information~\cite{supple}.

The advantage of targeted augmentation over random counterpart becomes evident in the presence of noise in the test trajectory data, which stems from the measurement error present in the experimental data. We simulate this effect by generating the test dataset from the models and then applying Gaussian noise to every point of the trajectories. Since every trajectory has been rescaled so that the standard deviation of particle displacement per time step is unity, the amplitude of the Gaussian noise can be regarded as the inverse signal-to-noise ratio (SNR). In Fig.~\ref{figure5}, we show that targeted augmentation leads to more robust performance than random augmentation as the noise becomes stronger by increasing $1/\mathrm{SNR}$. This demonstrates that Grad-CAM indeed captures the features that contain information about the underlying diffusion mechanism, which can be utilized to mitigate the adverse effects of data contamination by erasure or noise.

\begin{figure*}
\centerline{\includegraphics[width=\linewidth]{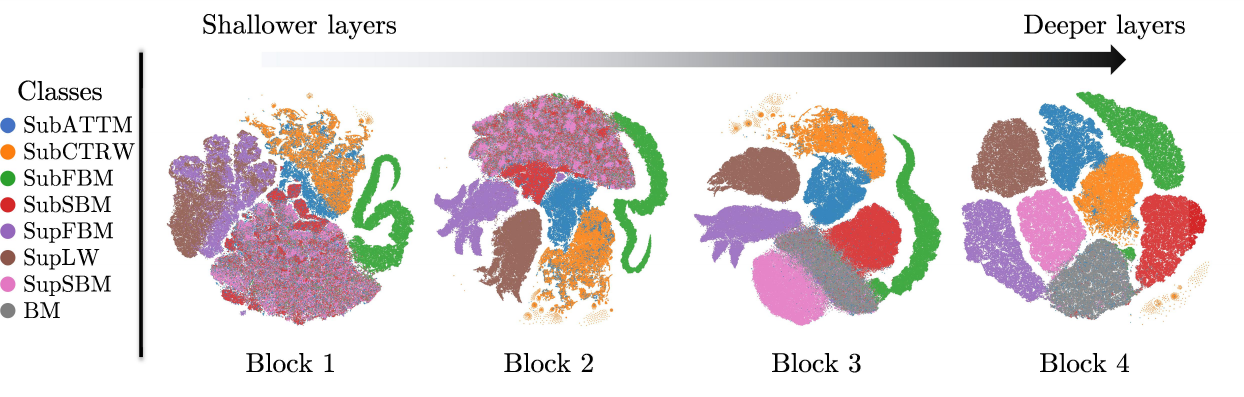}}
\caption{Visualization of the classification process across the four convolutional blocks of ResAnDi architecture. The high-dimensional output of each convolutional block is embedded into a two-dimensional space using t-SNE. Each scatter plot includes $400,000$ trajectories of the training dataset, with every point corresponding to a single trajectory.}
\label{figure6}
\end{figure*}

\section{Features indicated by Grad-CAM} \label{sec:statistics}

Now that the relevance of Grad-CAM to the classification of anomalous diffusion mechanisms has been demonstrated, we address the question of which characteristics of the mechanisms are captured by the Grad-CAM score. We take a two-step approach. First, by visualizing how ResAnDi architecture encodes trajectories generated by different mechanisms at various depths of the network, we construct a set of statistical features which might be useful for the classification task. Second, we calculate the correlations between these features and the Grad-CAM score, quantifying the extent to which Grad-CAM highlights those characteristics. For the ease of statistical analysis, through this section we use a newly trained ResAnDi, which was trained on a noiseless dataset comprising $8\times10^4 \times 5$ trajectories of temporal duration fixed at $1000$. For the calculation of correlations, a test dataset of the same size is used.

\subsection{Visualization of the classification process}

As described in Sec.~\ref{ssec:deep_learning}, ResAnDi architecture comprises four convolutional blocks, which successively process the trajectory data to identify its underlying diffusion mechanism. For every input trajectory, each block yields a high-dimensional output vector. Since ResAnDi is trained to distinguish between trajectories generated by different mechanisms, the output vectors must be clustered so that trajectories of ``similar'' mechanisms are close to each other, while those from ``dissimilar'' mechanisms are farther apart. In shallower layers, trajectories are likely to be clustered according to local features. In deeper layers, the network focuses more upon longer-range features. Hence, examining clustering patterns across different layers can provide us with useful clues as to which statistical features are used by ResAnDi to classify trajectories.

In Fig.~\ref{figure6}, we visualize how each convolutional block clusters the training dataset trajectories by embedding the output vectors in two dimensions via t-distributed stochastic neighbor embedding (t-SNE). In the output of Block~1, we can discern four mechanism clusters, namely [SubFBM], [SupFBM, SupLW], [SubATTM, SubCTRW], and [SubATTM, SubSBM, SupSBM, BM] (note that SubATTM appears in the intersection between two different clusters). Among these, SupFBM, SupLW, SubATTM, and SubCTRW are identified as individual mechanisms in the output of Block~2. Finally, the [SubSBM, SupSBM, BM] cluster is fully classified only after Block~3, indicating that long-range features are required to distinguish between these mechanisms. Based on these observations, we propose four statistics that may be utilized by ResAnDi for the classification task.

\subsection{Statistics correlated with Grad-CAM}
\label{ssec:statistics}

To facilitate further statistical analysis, we begin with partitioning each trajectory of the training dataset into subtrajectories, so that the Grad-CAM score can be assigned to each of them according to a procedure similar to the one described in Sec.~\ref{ssec:grad_cam}. However, in this case, we let the subtrajectories overlap with each other, so that each subtrajectory is long enough for reliable statistics. See the Supplementary Information~\cite{supple} for details.


Now, guided by Fig.~\ref{figure6}, let us construct the statistics that would be relevant to the classification task. In the output of Block~1, the SubFBM stands out as a single clearly distinct mechanism. Noting that SubFBM is the only model producing negative correlations between consecutive particle displacements, it is natural to conjecture that Block~1 utilizes the \textit{Autocorrelation} (AC) defined as
\begin{align}
	\mathrm{AC} \equiv \frac{1}{2}\sum_{r\in\{x,\,y\}}\frac{\langle\Delta r_t\,\Delta r_{t+1}\rangle-\langle \Delta r_t \rangle^2}{\llangle \Delta r_t^2 \rrangle}.
\end{align}

Here $\Delta r_t$ represents the displacement in the $x$ or $y$ direction during the time interval from $t$ to $t+1$, and $\llangle X^n\rrangle$ denotes the $n$th cumulant of observable $X$ over a chosen subtrajectory, with $\langle X \rangle \equiv \llangle X \rrangle$.

With AC thus identified, it seems natural that SupFBM and SupLW should be clustered together by Block~1 since they share positive AC. But while SupLW tends to make the particle move in the same direction for a prolonged period of time, SupFBM results in frequent changes in the direction of motion. Thus, to distinguish these two mechanisms, we expect that Block~2 takes advantage of \textit{Consistency} (CS) defined as
\begin{align}
    \mathrm{CS} \equiv \mathrm{AC} \times \overline{\sqrt{\llangle(\Delta \theta_t/\pi)^2\rrangle}},
\end{align}
where
\begin{align}
    \Delta\theta_t \equiv \arccos \frac{\Delta x_{t-1}\Delta x_t + \Delta y_{t-1}\Delta y_t}{\sqrt{(\Delta x_{t-1}^2+\Delta y_{t-1}^2)(\Delta x_t^2+\Delta y_t^2)}}
\end{align}
is the change in the direction of motion occurring at time $t$, and the horizontal line indicates that the quantity is min-max normalized to the interval $[0,\,1]$. Clearly, high (low) CS corresponds to SupFBM (SupLW).

As for the [SubATTM, SubSBM, SupSBM, BM] cluster, the component mechanisms are all characterized by locally Gaussian displacements. This motivates us to divide each subtrajectory into $n$ subintervals and consider the \textit{Non-Gaussianity} (NG) defined as
\begin{align}
	\mathrm{NG} \equiv \overline{\left|\frac{1}{2n}\sum_{r\in\{x,\,y\}}\sum_{i=1}^n \frac{\llangle \Delta r_t^4 \rrangle_i}{\llangle \Delta r_t^2 \rrangle_i^2}\right|},
\end{align}
where $\llangle \cdot \rrangle_i$ denotes a cumulant calculated over the $i$th subinterval. Since the fourth cumulant vanishes for the Gaussian distribution, NG is greater if the subtrajectory deviates farther from the Gaussian statistics at the subinterval level.

Now, we can guess how SubATTM and SubCTRW are established as individual mechanisms after Block~2. Both are distinct from the [SubSBM, SupSBM, BM] cluster in that their trajectories feature sudden changes in the magnitude of displacements. They are also distinguished from each other by the locally Gaussian nature of SubATTM and the strong non-Gaussianity of SubCTRW. Thus, we are led to consider \textit{Singularity} (SG) defined as
\begin{align}
	\mathrm{SG} \equiv \overline{\max_{r\in\{x,\,y\}}\!\left[\left\llangle\left(\frac{\Delta r_{t+1}+\epsilon}{\Delta r_{t}+\epsilon}\right)^2\right\rrangle^{1/2}\right]} \times \mathrm{NG},
\end{align}
where $\epsilon$ is a small positive number introduced to prevent the quantity from diverging. Throughout this study, we use $\epsilon = 10^{-6}$.
We note that high (low) SG indicates SubATTM (SubCTRW).

Finally, there still remains the [SubSBM, SupSBM, BM] cluster, which persists up to Block~3. The three mechanisms are similar in that their fluctuations are locally Gaussian without any sudden changes in the magnitude of displacements. Their unique characteristics become apparent only when one observes how diffusivity gradually changes over time. As a measure of this property, we propose \textit{Varying Diffusivity} (VD) defined as
\begin{align}
	\mathrm{VD} \equiv \frac{1}{2n}\sum_{r\in\{x,\,y\}}\frac{\llangle \Delta r_t^2 \rrangle_n^{1/2} - \llangle \Delta r_t^2 \rrangle_1^{1/2}}{\llangle \Delta r_t^2 \rrangle^{1/2}} \times \mathrm{NG},
\end{align}
which quantifies how diffusivity changes over each subtrajectory. We note that positive (negative) VD indicates SupSBM (SubSBM).

\begin{figure*}
\centerline{\includegraphics[width=\linewidth]{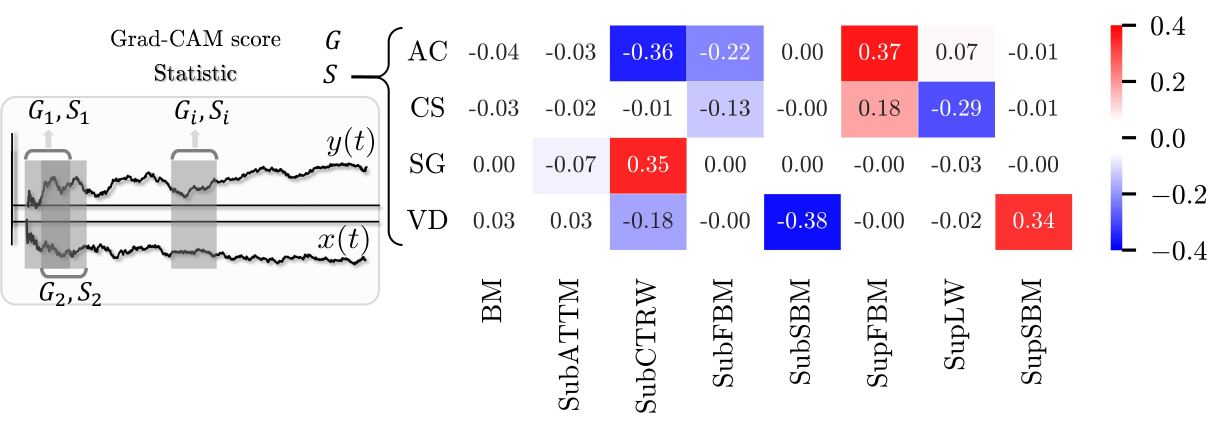}}
\caption{Pearson correlation coefficients between the Grad-CAM score and each of the four statistics constructed in Sec.~\ref{ssec:statistics}. Note that the Grad-CAM score and the statistics are assigned to every overlapping subtrajectories, as illustrated in the inset. Correlations are obtained using a test dataset consisting of 400,000 trajectories.
}
\label{figure7}
\end{figure*}

In Fig.~\ref{figure7}, we show how the statistics constructed by the above procedure correlate with the Grad-CAM score of the subtrajectories. The results can be interpreted as follows.

\begin{itemize}
    \item AC, the most local measure, is negatively correlated with the Grad-CAM scores of SubFBM and SubCTRW. The reason for the former is clear, for SubFBM features negative AC throughout the trajectories. The latter may stem from the ResAnDi assigning high Grad-CAM scores to subtrajectories with abrupt jumps, which also induces negative AC for those subtrajectories. Meanwhile, AC exhibits positive correlation with the Grad-CAM scores of SupFBM and SupLW, while correlations with the other mechanisms are largely negligible. This reflects how Block~$1$ clusters the diffusion mechanisms, clearly separating the [SubFBM] and the [SupFBM, SupLW] clusters from the rest.
    \item CS exhibits significant correlations of opposite signs with the Grad-CAM scores of SupFBM and SupLW. This is consistent with our conjecture that Block~2 utilizes the statistic to distinguish between the two mechanisms that commonly feature positive AC. We note that CS also exhibits negative correlation with the Grad-CAM score of SubFBM, which might have been inherited from the behaviors of AC.
    \item SG and the Grad-CAM score exhibit a significant positive correlation for SubCTRW and a weak negative correlation for SubATTM. This seems to confirm that Block~2 indeed uses the statistic to separate the two mechanisms from the others.
    \item VD, the most nonlocal measure, exhibit significant correlations of opposite signs with the Grad-CAM scores of SupSBM and SubSBM, suggesting that the statistic is indeed used by Block~3 to classify the two mechanisms. We also note that the Grad-CAM score of SubCTRW also exhibits a negative correlation with VD, which may stem from the strong non-Gaussianity of the SubCTRW trajectories.
\end{itemize}

In the end, Fig.~\ref{figure7} shows that every diffusion mechanism exhibits a distinct correlation profile with the four statistics introduced above, except for SubATTM and BM that are distinguishable only by the weak negative correlation shown by SG. Indeed, as shown in Fig.~S2 of the Supplementary Information~\cite{supple}, ResAnDi finds it difficult to distinguish SubATTM from BM. These results demonstrate how the Grad-CAM scores can provide some insight into which statistical features are utilized by deep learning to decode the underlying diffusion mechanism of a trajectory.

\section{Summary and outlook} \label{sec:summary}

In this study, we utilized Grad-CAM, a technique developed for explainable AI, to highlight which parts of particle trajectories are crucial for the deep learning algorithm to identify the underlying diffusion mechanism. By observing how targeted erasure of trajectories based on Grad-CAM impairs the machine learning performance, we found that Grad-CAM indeed captures the informative parts of the trajectories, which can also be utilized to enhance the dataset augmentation method. Furthermore, by measuring correlations between the Grad-CAM score and trajectory statistics of varying nonlocality, we could elucidate the process through which deep learning differentiates between different diffusion mechanisms, step by step.


Our results have twofold implications. First, they demonstrate that Grad-CAM provides a useful measure of which parts of the training dataset are more informative than the rest. This suggests that one may design an active learning algorithm that makes best use of the available trajectory dataset by incorporating Grad-CAM into the training procedure. Second, our results confirm the intuition that deep learning decodes the trajectory data by first focusing on local features and then gradually broadening the scope to nonlocal features. Designing a statistical inference method inspired by such multiscale attention implemented by deep learning would be a worthwhile direction of future research.




\begin{acknowledgments}
This study was supported by the Basic Science Research Program through the National Research Foundation of Korea (NRF Grant No. 2022R1A2B5B02001752) and the Global-LAMP Program of the National Research Foundation of Korea (NRF) grant funded by the Ministry of Education (No. RS-2023-00301976).
\end{acknowledgments}

\bibliographystyle{apsrev4-2}
\bibliography{ref}


\pagebreak
\pagebreak
\widetext

\begin{center}
\textbf{\large Supplementary Information:\\Exploring how deep learning decodes anomalous diffusion  via Grad-CAM}
\end{center}

\setcounter{equation}{0}
\setcounter{figure}{0}
\setcounter{table}{0}
\setcounter{page}{1}
\setcounter{section}{0}
\renewcommand\thesection{\Alph{section}}
\renewcommand\thesubsection{\arabic{subsection}}
\makeatletter
\renewcommand{\theequation}{S\arabic{equation}}
\renewcommand{\thefigure}{S\arabic{figure}}

All the algorithms used in our study can be found in the accompanying code repository on GitHub, available at \url{https://github.com/peardragon/ResAnDi}.

\section{Anomalous Diffusion Trajectory Dataset}\label{apped.A} 

Theoretical models of anomalous diffusion (whose exponent is denoted by $\alpha$) used in this study are as follows:
\begin{itemize}
    \item Annealed Transient Time Motion (ATTM): The particle exhibits the Brownian motion with its diffusion coefficient $D$ varying over time~\cite{ATTM1}. When $D$ changes, the new value is randomly chosen from the distribution $P(D) \sim D^{\sigma-1}$, where $D \leq 1$ and $\sigma \in (0,3]$. The chosen value of $D$ is maintained for the time interval $\Delta t = D^{-\gamma}$, where $\sigma < \gamma < \sigma + 1$. Then, the particle exhibits subdiffusion whose exponent is given by $\alpha = \sigma/\gamma$. Examples of systems showing this behavior include proteins subject to receptor-ligand interactions~\cite{ATTMapplication}.
    
    \item Continuous Time Random Walk (CTRW): The model is used to describe a particle moving in a landscape riddled with potential wells of various depths~\cite{CTRW1}. When the particle is trapped in a well, it remains static for the waiting time distributed as $\psi(\tau) \sim \tau^{-1-\alpha}$, where $0 < \alpha < 1$. Then it instantaneously moves to a nearby trap, whose distance from the previous trap follows the normal distribution $\Delta x \sim \mathcal{N}(0, D)$.
    
    \item Fractional Brownian Motion (FBM): The motion of the particle is driven by a Gaussian noise whose correlation satisfies
    \begin{align}
        \langle\xi_i(t_1) \xi_j(t_2)\rangle= K \alpha(\alpha-1)\left|t_1-t_2\right|^{\alpha-2} \delta_{i j} + 2 K \alpha\left|t_1-t_2\right|^{\alpha-1} \delta\left(t_1-t_2\right) \delta_{i j},
    \end{align}
    where $K > 0$ and $0 < \alpha < 2$. This model is commonly applied to particles moving in viscoelastic media~\cite{FBMapplication}. Note that the model reduces to the ordinary Brownian motion when $\alpha = 1$.
    
    \item L\'{e}vy walks (LW): The particle exhibits ballistic motion punctuated by random switching of directions. The speed of the particle after each switching is randomly chosen in the interval $v \in [-10, 0) \cup (0,10]$, and the waiting time between the switchings is distributed as $\psi(t) \sim t^{-1-\sigma}$. Depending on the value of $\sigma$, the diffusion exponent $\alpha$ is given by $\alpha = 2$ when $0 < \sigma < 1$ and $\alpha = 3 - \sigma$ when $1 < \sigma < 2$. The model is commonly applied to describe hunting and gathering strategies of animals~\cite{LW1}.
    
    \item Scaled Brownian Motion (SBM): The particle exhibits the Brownian motion whose diffusion coefficient changes in time according to $D(t) = \alpha D t^{\alpha -1}$, where $0 < \alpha < 2$~\cite{SBM1}. The model is used to describe phenomena such as Fluorescence Recovery After Photobleaching (FRAP)~\cite{SBMapplication}.
\end{itemize}

\section{Neural network architecture and training procedure} 

\begin{figure*}[!ht]
\centerline{\includegraphics[width=0.8\textwidth]{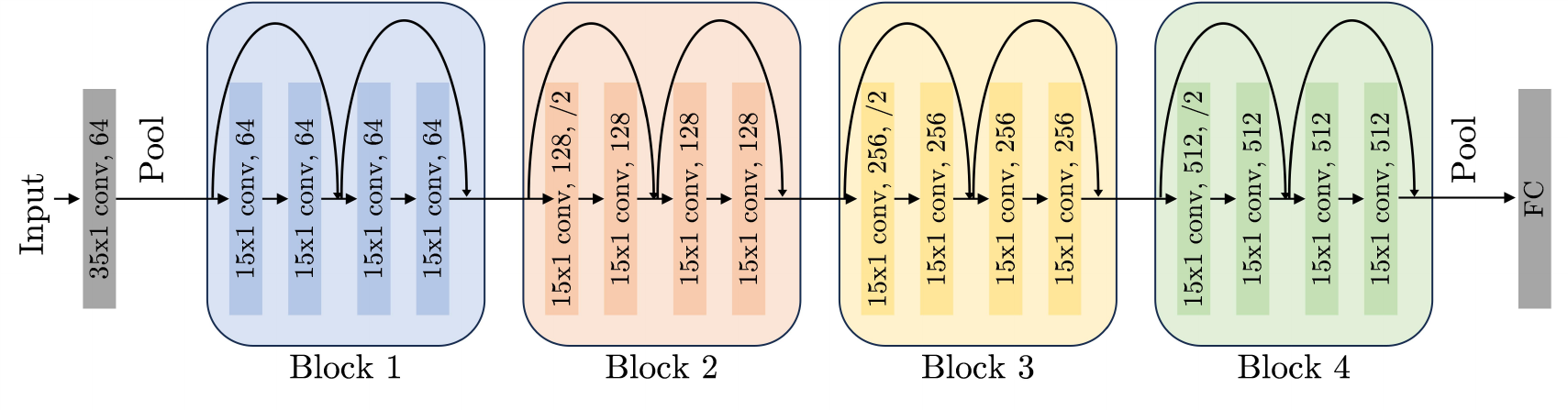}}
\caption{
Structure of ResAnDi, which consists of 17 convolutional layers and a fully-connected (FC) layer. ``Pool'' indicates a pooling operation, and shortcut connections are indicated by curved arrows. For each convolutional layer, ``$a$x$b$ conv, $n$" means the filter size of $a$x$b$ and the number of output channels given by $n$. Moreover, ``/2" is used to signify the stride.
}
\label{supfig0}
\end{figure*}

\subsection{ResAnDi architecture} 
The neural network architecture of ResAnDi (see Fig.~\ref{supfig0}) is based on ResNet18~\cite{resnet}, which has been adapted to suit the input dimensions of $C'\times H'\times W' = 2\times1\times1000$. The modifications are summarized in Table ~\ref{suptab1}.

\begin{table*}[!h]
\caption{Comparison between ResNet18 and ResAnDi}
\label{suptab1}
\begin{tabular}{l|ll|ll}
\hline
Input shape & \multicolumn{2}{l|}{3 $\times$ 225 $\times$ 225}                               & \multicolumn{2}{l}{2 $\times$ 1000 $\times$ 1}                                      \\ \hline
Layer name  & \multicolumn{1}{l|}{Output size (ResNet18)}           & Filter map (ResNet18)  & \multicolumn{1}{l|}{Output size (ResAnDi)}           & Filter map (ResAnDi) \\ \hline
$a$x$b$ conv, 64       & \multicolumn{1}{l|}{64 $\times$ 112 $\times$ 112}     & 7 $\times$ 7           & \multicolumn{1}{l|}{64 $\times$ 500 $\times$ 1}      & 35 $\times$ 1                 \\ \hline
Block 1   & \multicolumn{1}{l|}{64 $\times$ 56 $\times$ 56}       & 3 $\times$ 3           & \multicolumn{1}{l|}{64 $\times$ 250 $\times$ 1}      & 15 $\times$ 1                 \\ \hline
Block 2    & \multicolumn{1}{l|}{128 $\times$ 28 $\times$ 28}                   & 3 $\times$ 3           & \multicolumn{1}{l|}{128 $\times$ 125 $\times$ 1}     & 15 $\times$ 1                 \\ \hline
Block 3    & \multicolumn{1}{l|}{256 $\times$ 14 $\times$ 14}                   & 3 $\times$ 3           & \multicolumn{1}{l|}{256 $\times$ 63 $\times$ 1}      & 15 $\times$ 1                 \\ \hline
Block 4   & \multicolumn{1}{l|}{512 $\times$ 7 $\times$ 7}                     & 3 $\times$ 3           & \multicolumn{1}{l|}{512 $\times$ 32 $\times$ 1}      & 15 $\times$ 1                 \\ \hline
\end{tabular}
\end{table*}

\subsection{Training with trajectories of varying lengths}

For training, we used trajectories with temporal durations (lengths) ranging from 10 to 1000. Each diffusion mechanism generated $5\times10^4$ trajectories, so the entire training dataset consisted of $8\times5\times10^4$ trajectories. Meanwhile, the validation set contained $8\times10^{4}$ trajectories.

Before using the trajectories as input, they went through preprocessing composed of two steps. First, we applied simple zero-padding to fix the input trajectory length to $1000$. Specifically, zero-padding was added before the beginning of the trajectory. Next, we min-max normalized the position values of the trajectories to the interval [0,1]. The preprocessed trajectory $\overline{r}(t) \in \{\overline{x}(t), \overline{y}(t)\}$ was derived by normalizing the original trajectory $r(t)\in{x(t), y(t)}$ according to $\overline{r}(t) = ({r(t)-r_{\text{min}}})/({r_{\text{max}}-r_{\text{min}}})$, where $r_{\text{min}}$ and $r_{\text{max}}$ are the minimum and maximum values of $r(t)$, respectively.

These preprocessed trajectory datasets were then put into ResAnDi for training. Each training session used the Adam optimizer from PyTorch, with a learning rate of $\gamma = 0.0001$ and a batch size of 64. To prevent overfitting, Early Stopping was employed, terminating the training if the cross-entropy loss function did not improve for 10 consecutive iterations on the validation dataset.
Additionally, we implemented a Step Learning Rate Scheduler, which halved the learning rate every 10 iterations.

\subsection{Training with augmented datasets}

In addition to the original dataset of $8\times10^4$ trajectories, $8\times10^4\times0.6$ trajectories were added to construct an augmented dataset. For targeted augmentation, trajectories whose Grad-CAM scores were above the $60$th percentile were included in the added dataset. The rest of the training process remained the same as the previous cases, utilizing the Adam optimizer, Early Stopping, and a learning rate scheduler with an initial learning rate of $0.0001$. Finally, the accuracy is measured using the validation set of $8\times10^4$ trajectories, averaging over $5$ model outcomes for each data point.

\subsection{Training with fixed-length trajectories}

The results of Sec.~IV were obtained using noiseless trajectories of temporal duration fixed at $1000$. For this case, ResAnDi was trained using $8\times5\times10^4$ trajectories, and the validation set of $8\times10^4$ trajectories was used to evaluate the performance. The remaining aspects of the training process were the same as before, utilizing the Adam optimizer, Early Stopping, and a learning rate scheduler with an initial learning rate of $0.0001$.

\section{Physical interpretation of classification results}

\begin{figure}[!ht]
\centerline{\includegraphics[width=0.8\linewidth]{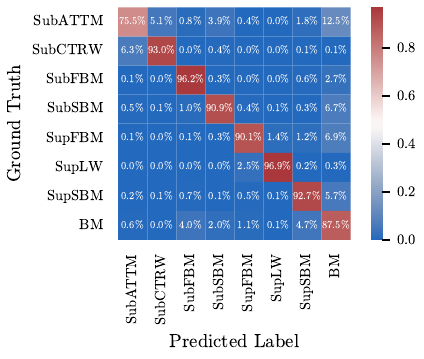}}
\caption{Confusion matrix showing the classification performance of ResAnDi. The colors indicate the probabilities.}
\label{supfig1}
\end{figure}

ResAnDi achieved an overall classification accuracy of $90.36 \%$ over the validation dataset $8 \times 10^4$ trajectories, which consisted of $10^4$ trajectories generated by each of the $8$ diffusion mechanisms. The classification results for each class are shown by a confusion matrix in Fig~\ref{supfig1}.

\begin{figure*}[!ht]
\centerline{\includegraphics[width=\linewidth]{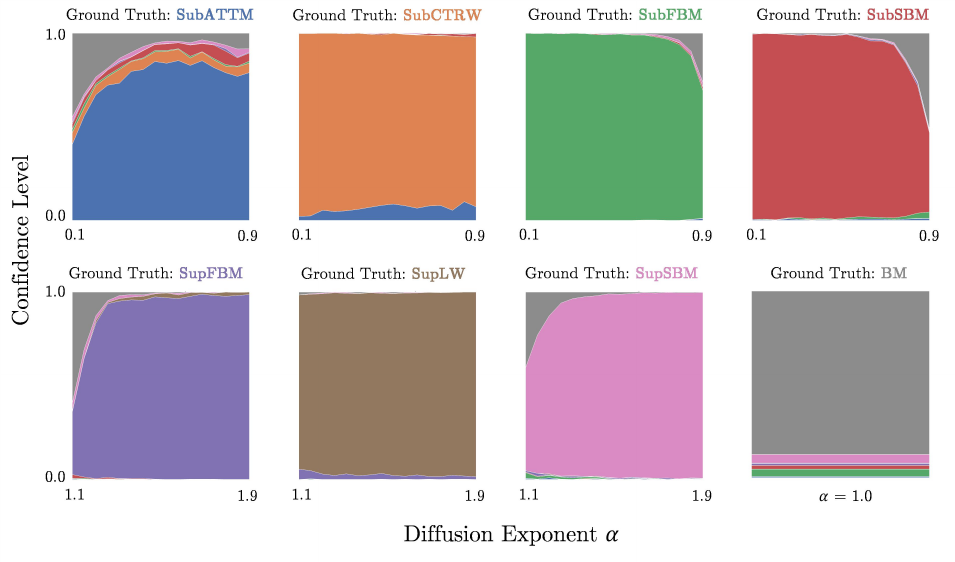}}
\caption{Mean predicted confidence levels of various diffusion models for the given ground-truth model and the diffusion exponent $\alpha$. The mean confidence levels were computed by averaging over $10^4$ trajectories for each diffusion model. Note that the range of $\alpha$ is given by $[0.1,0.9]$ ($[1.1,1.9]$) for the models exhibiting subdiffusion (superdiffusion). As for BM, the only possible value of $\alpha$ is $1$ by definition.}
\label{supfig2}
\end{figure*}

While the results are highly accurate overall, details of Fig.~\ref{supfig1} reveal common misclassification patterns. The most significant source of error lies in the misclassification of trajectories as BM. It is also notable that ResAnDi tends to confuse SubATTM with SubCTRW.

To explore these properties further, in Fig.~\ref{supfig2} we analyze the classification results for different ground-truth models as the diffusion exponent $\alpha$ is varied. In the figure, the width of each colored region at a fixed value of $\alpha$ indicates the mean confidence level of the model (\textit{i.e.}, the proportion of trajectories classified as the model) represented by the same color.

Except for SupLW which remains accurately distinguishable throughout the whole range of $\alpha$, the confidence levels of all the other models change significantly as $\alpha$ is varied. For SubFBM, SubSBM, SupFBM, and SupSBM, accuracy tends to decrease as $\alpha$ approaches $1$, reflecting that most of the models described in \ref{apped.A} reduce to BM in the limit.

Interestingly, SubATTM shows the opposite trend: it is more likely to be misclassified as BM when $\alpha$ approaches $0$, while it is more easily confused with SubCTRW when $\alpha$ increases. This can be intuitively understood as follows. The generation mechanism of SubATTM described in \ref{apped.A} implies that, when $\alpha$ approaches $0$, so should $\sigma$, which suppresses the heterogeneity of the diffusion coefficient $D$ to the strongest extent. Due to this effect, it becomes more difficult to distinguish SubATTM from BM when $\alpha$ is closer to $0$. On the other hand, when $\alpha$ increases, $\sigma$ is also allowed to have greater positive values. While this renders SubATTM more distinguishable from BM, it also increases the likelihood that the SubATTM trajectories alternate between long intervals of tiny $D$ and short intervals of huge $D$, which may appear very similar to the SubCTRW trajectories that alternate between long static periods and instantaneous jumps. Thus, the chance of confusing SubATTM with SubCTRW increases as $\alpha$ moves away from $0$.

\section{Choice of the subtrajectory length with a single Grad-CAM value} 

In our study, we assigned a single Grad-CAM score to a subtrajectory of length $225$. For instance, a subtrajectory from $t=1$ to $t=225$ corresponds to a single Grad-CAM value, and a subtrajectory from $t=26$ to $t=250$ corresponds to the next Grad-CAM value, etc. This was based on our analysis of how each channel in the final convolutional layer responds to the local signals in the input, as explained below.

\begin{figure*}[!ht]
\centerline{\includegraphics[width=0.6\linewidth]{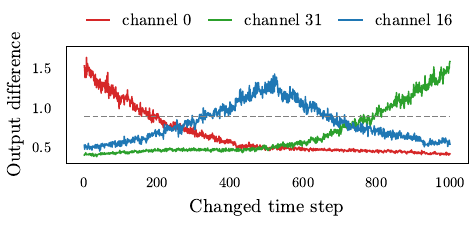}}
\caption{Change of the $0$th, $16$th, and $31$st channel output as the value of the input is changed from $0$ to $1$ at a single time step. The dashed horizontal line indicates the threshold $0.9$, which was used to determine the subtrajectory length.}
\label{response}
\end{figure*}

Let us consider an input time series that is zero at all time steps. Then we measure how much the output of one of the $32$ channels (to which a single Grad-CAM score is assigned, as described in Sec.~III.~A) in the final convolution layer changes if the value of the input changes from $0$ to $1$ at the $i$-th time step. The results are shown in Fig.~\ref{response}, which reveals that each channel (indicated by colors) is most sensitive only to a limited domain of the input. The subtrajectory length was chosen based on this observation.

\vspace*{\fill}
\pagebreak

\section{Example trajectories}

For comparison with discussions in Sec. IV, here we explicitly show how Grad-CAM highlights the characteristic portions of the particle trajectories generated by the diffusion models.

\begin{figure*}[!ht]
\centerline{\includegraphics[width=\textwidth]{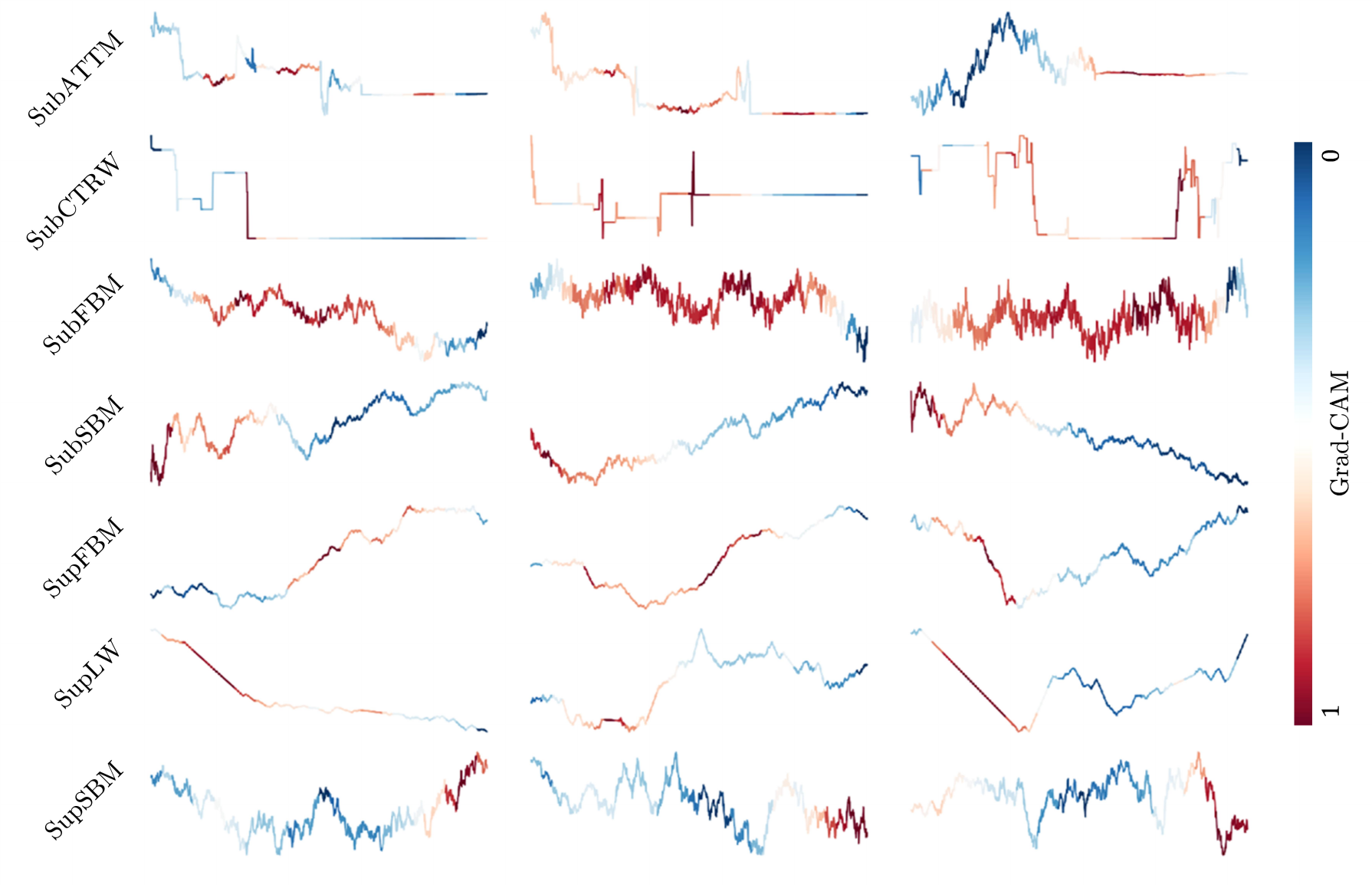}}
\caption{Sample trajectories of the diffusion models and their Grad-CAM score profiles. The vertical displacements indicate $x(t)$, and the Grad-CAM values are color-coded after min-max scaling.}
\label{supfig4}
\end{figure*}

For each diffusion model, Grad-CAM highlights the following aspects:
\begin{itemize}
\item SubATTM: Long periods of small diffusion coefficient tend to be highlighted.
\item SubCTRW: Regions with frequent jumps tend to be highlighted.
\item SubFBM: All non-boundary regions are highlighted, since negative correlations pervade the time series.
\item SubSBM: Only the initial portion of the trajectory is highlighted, where diffusivity rapidly decreases.
\item SupFBM: Regions exhibiting persistent increases or decreases tend to be highlighted.
\item SupLW: The longest piece of ballistic motion tends to be highlighted.
\item SupSBM: Only the terminal portion of the trajectory is highlighted, where diffusivity rapidly increases.
\end{itemize}


\end{document}